\documentclass[letterpaper, 10 pt, journal, twoside]{IEEEtran}

\IEEEoverridecommandlockouts                              

\usepackage{graphics} 
\usepackage{epsfig} 
\usepackage{times} 
\usepackage{amsmath} 
\usepackage{amssymb} 
\usepackage{xcolor}
\usepackage{booktabs}
\usepackage{multirow}
\usepackage{graphicx}
\usepackage{enumitem}
\usepackage{url}
\newcommand{\eg}{\emph{e.g.},}

\newcommand{\etal}{\emph{et~al.}}
\usepackage{epstopdf}

\definecolor{pastelgreen}{RGB}{25,190,0}

\newcommand{\ccyan}[1]{{\color{black}#1}}

\usepackage{pgfplotstable}
\usepackage[table]{xcolor}
\usepackage{booktabs}

\definecolor{bestRed}{HTML}{F8CECC}   
\definecolor{worstYellow}{HTML}{FFFFBF} 

\newcommand{\cc}[1]{\cellcolor{bestRed!#1!worstYellow}}

\title{\LARGE \bf NEO: NeRF It Once, Edit It Many Times 
\\ for Continuous Object Manipulation
}

\author{Mikołaj Zieli\'nski$^{1,2}$, David Hall$^{2}$, Dominik Belter$^{1}$, Peyman Moghadam$^{2}$
\thanks{Manuscript submitted: February, 22, 2026}
\thanks{* Mikołaj Zieliński and Dominik Belter were supported by the National Science Centre, Poland, under research project no UMO-2023/51/B/ST6/01646.}
\thanks{$^{1}$Institute of Robotics
and Machine Intelligence, Poznan University of Technology, 61-131 Poznań,
Poland. E-mail: name.surname@put.poznan.pl}%
\thanks{$^{2}$CSIRO Robotics, CSIRO, Australia. E-mail: firstname.lastname@csiro.au}%
}

\begin{document}

\bstctlcite{IEEEexample:BSTcontrol}

\def\our{NEO}

\maketitle

\begin{abstract}
    In this paper, we present \our{}, a unified framework providing language-guided NeRF editing for robotic manipulation.
Our paper introduces (i) a language-guided object removal that combines neural field resampling with multiview-consistent progressive inpainting, (ii) a direct NeRF weight editing method utilizing knowledge distillation, composing original and edited NeRFs via a teacher–student model, enabling coherent modeling of future scene states before a robot executes an action, and (iii) the first benchmark (\textit{NEO-Dataset}) for quantitatively evaluating NeRF scene editing methods suitable for robot manipulation.
We show that our approach outperforms state-of-the-art baselines in scene editing tasks, including object removal and pick-and-place robotic experiments, yielding visually coherent and geometrically consistent edits that reduce artifacts commonly introduced by prior methods.
Finally, we showcase the capability of \our{} for multi-stage robotic assembly tasks by preserving a persistent NeRF plus language-field representation after each edit, enabling iterative future-state scene representation prediction without requiring additional scene re-scanning.
\end{abstract}


\IEEEPARstart{N}{eural} Radiance Fields (NeRFs) have emerged as a powerful implicit neural scene representation, enabling high-fidelity novel view synthesis and dense 3D reconstruction from a set of images \cite{mildenhall2020nerf}.

Beyond computer vision, these methods are increasingly being explored in robotics \cite{wang2025nerfsroboticssurvey}, with applications in 3D scene reconstruction \cite{wang2023urbannerf,tao2024silvr}, navigation \cite{byravan2023nerf2real}, grasping~\cite{Song2025,rashid2023lerftogo}, segmentation \cite{engelmann2024opennerf}, and robotic manipulation \cite{rashid2023lerftogo,tseng2022clanerf,ichnowski2022dexnerf}.

A key advantage of NeRF-based representations for grasping and manipulation is that they encode 3D environments as continuous functions \cite{mildenhall2020nerf}, providing view-consistent geometry and appearance that can be queried at arbitrary resolutions. This enables NeRFs to provide direct continuous occupancy queries through a single network evaluation, which is advantageous for grasp synthesis and collision checking. Additionally, NeRFs typically provide accurate and globally consistent 3D reconstructions \cite{Fang2025}. This makes NeRFs appealing for tasks that require fine-grained spatial reasoning, such as selecting grasp points~\cite{Song2025,rashid2023lerftogo}, validating approach trajectories, teleoperation~\cite{Wilder-Smith2024_teleoperation}, and predicting occlusions during interaction~\cite{zhu2023occlusion}. NeRFs can also encode additional modalities such as language features \cite{kerr2023lerf}, enabling language-guided grasping and manipulation \cite{rashid2023lerftogo, shendistilled}. However, robotic manipulation is inherently an interactive process. Each action can change object poses, introduce new occlusions, or reveal previously unseen surfaces, which can invalidate the original NeRF. Consequently, many NeRF-based robot manipulation pipelines require re-observing or re-scanning the scene to update the representation before subsequent actions \cite{lu2024fast}, which is time-consuming and often impractical for real-world robotic applications.

An alternative approach is to edit the NeRF representation directly, enabling the scene model to be updated after interaction without re-scanning or re-optimizing from scratch. Several approaches have been proposed in this direction ~\cite{weber2023nerfiller,bartrum2024replaceanything3d,Wang2023_SEAL3D,chen2023neuraleditor,yuan2022nerfediting,wang2023learning}.
While these methods demonstrate compelling visual results, they often fall short of the robustness and consistency required for real-world robot manipulation.
In particular, edits should be permanent to support future scene updates \cite{yuan2022nerfediting,hausler2024reg}; removed objects must reveal consistent underlying geometry \cite{Wang2023_SEAL3D, wang2023learning}; the edited representation should remain reliable across viewpoints and subsequent actions \cite{Weder2023Removing}; and the editing process should avoid hallucinations \cite{bartrum2024replaceanything3d} \cite{rombach2022high}. These gaps highlight a pressing need for NeRF editing methods tailored to robotic tasks: methods that can update scenes reliably and permanently, while avoiding re-scanning after each robot action.

\begin{figure}[t]
    \centering
    \includegraphics[width=1.0\columnwidth]{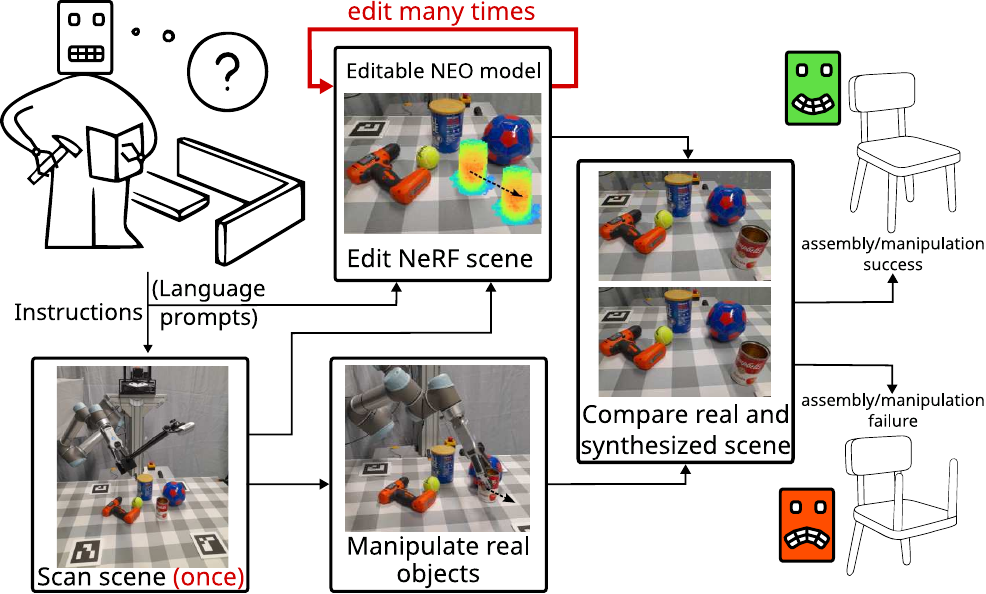}
    \vspace{-6mm}
    \caption{Example assembly task: given language prompts and a single initial scan, the robot executes multiple manipulations. \our{} updates the NeRF scene model after each interaction to predict the new scene state without re-scanning.}
    \label{fig:scenario}
\end{figure}

\begin{figure*}[t]
    \centering
    \includegraphics[width=0.99\textwidth]{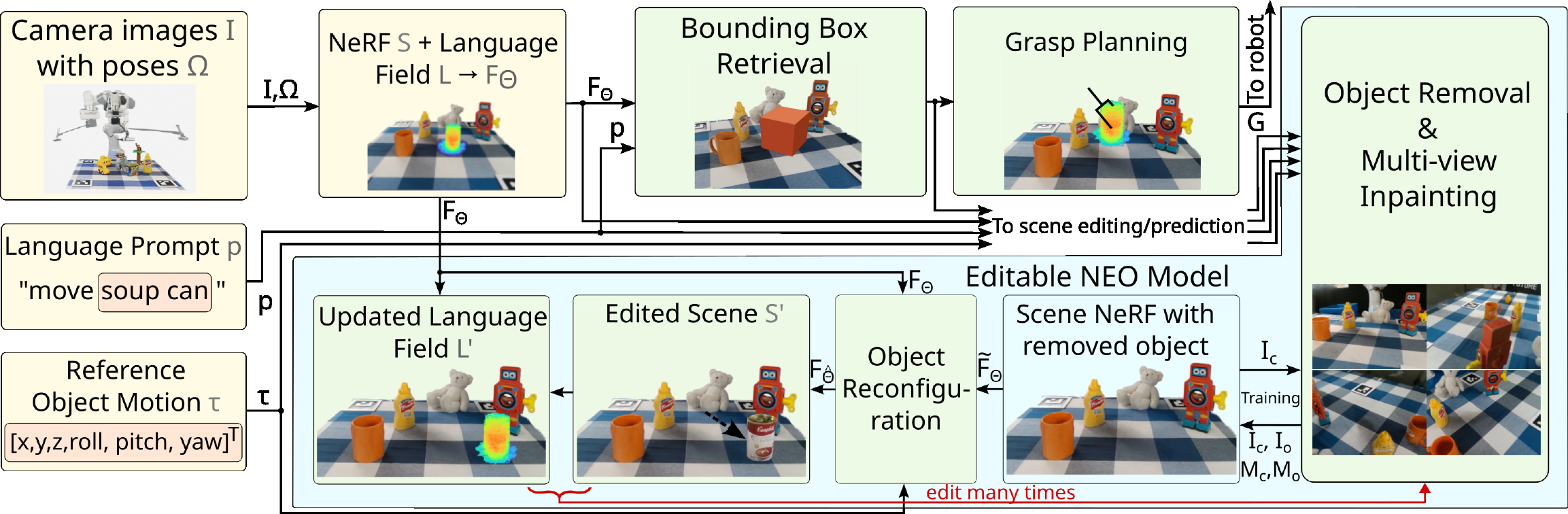}
    \vspace{-2mm}
    \caption{Architecture of the proposed editable \our{} model and its role within the robotic manipulation system. A set of camera images is used to build the NeRF and the language field. Given a language prompt, the object to be removed from the scene is selected. Prior to grasp planning and motion execution, the proposed editable \our{} model is used to reposition the object within the neural field and to predict the scene state after the robotic interaction. This is achieved through the proposed object removal, multiview inpainting, and object reconfiguration modules.}
    \label{fig:architecture}
    \vspace{-2mm}
\end{figure*}

To address these challenges, we present \textbf{\our}, a NeRF editing technique specifically designed for language-guided robotic manipulation that enables geometrically consistent scene modifications without requiring additional re-scanning. Given a text prompt specifying the object to be manipulated, our method performs a three-stage editing pipeline: (i) the object is removed from the scene representation using adaptive ray re-sampling method and image rendering; (ii) multiview-consistent inpainting reconstructs the occluded regions to ensure geometric plausibility; and (iii) the object is reinserted at a new location using object-centered knowledge distillation, preserving it permanently in the updated scene.
All edits are applied directly to the NeRF weights, making them persistent across consecutive scene modifications. We also support continued language-guided interactions involving the same object across subsequent actions.
In addition, \textbf{\our} supports batch edits and fine-grained adjustments, such as relocating objects within small spatial offsets.
This framework enables prediction and visualization of the edited scene prior to execution, allowing the robot to anticipate outcomes before acting.
The contributions of this work are:
\begin{itemize}[leftmargin=*]
\item A unified NeRF editing framework for robotic manipulation that directly updates network weights to enable realistic, geometrically consistent, and persistent scene modifications. The pipeline includes object removal via neural field resampling, multiview-consistent progressive inpainting for plausible reconstruction of revealed regions, and supports continuous, language-guided interaction without requiring additional scene re-scanning.
\item A scene-focused knowledge distillation method that reconfigures scenes by composing an originally trained NeRF with an edited NeRF, using a teacher–student model to permanently preserve scene changes.
\item The \textit{NEO-Dataset} benchmark for NeRF editing in robotic manipulation scenarios, with registered pre- and post-edit scenes to enable rigorous, quantitative evaluation.  

\end{itemize}

\section{LITERATURE REVIEW}

\subsection{NeRF Editing}

Several recent NeRF-based scene editing methods address object manipulation and appearance editing; however, challenges remain in object removal and in maintaining geometrically consistent 3D reconstructions, particularly under sequential edits. The Distilled Feature Fields (DFF) method~\cite{kobayashi2022decomposing} removes content by suppressing densities for ray samples within a user-specified region. Because the sampling strategy is not adapted after removal, regions behind the edited volume can become under-sampled, which often reduces reconstruction fidelity. In addition, DFF does not explicitly handle surfaces that were never observed in the original views, leaving newly revealed regions unconstrained.

Reconstructing such unobserved regions can be approached by applying 2D inpainting independently to each view; however, this often leads to view-inconsistent completions and degraded 3D reconstructions~\cite{Weder2023Removing}. NeRFiller~\cite{weber2023nerfiller} mitigates this issue by using a generative inpainting model to complete missing regions and then distilling the completed views into a NeRF, improving cross-view consistency. Nevertheless, this process typically requires retraining after object removal and can still introduce artifacts in the edited regions.

Several other methods explore related directions but do not provide the functionality needed to model the outcomes of robotic manipulation. RAM3D~\cite{bartrum2024replaceanything3d} performs inpainting after object removal but does not support object relocation. NeuralEditor~\cite{chen2023neuraleditor} edits derived point-cloud representations rather than directly updating a NeRF, while NeRF-Editing~\cite{yuan2022nerfediting} focuses on appearance deformation through ray bending and does not address object removal or rigid 3D transformations. Methods that explicitly separate object and background representations~\cite{wang2023learning} still rely on relatively limited inpainting, which can yield visual inconsistencies. DATENeRF~\cite{Rojas2024_DATENeRF} and ViCA-NeRF~\cite{Dong2023_ViCA-NeRF} improve multiview consistency for appearance edits, but are not designed for object removal or manipulation. Seal-3D~\cite{Wang2023_SEAL3D} introduces a teacher--student framework for scene editing, but does not address object removal, which is essential for manipulation scenarios.

\subsection{Scene Editing and Prediction for Robotic Interaction}

Several approaches have explored scene editing in the context of robotic interaction. Abou-Chakra \etal{}~\cite{Abou-Chakra2024} employ Gaussian Splatting to initialize particle-based simulations and incorporate correction mechanisms for closed-loop interaction, enabling iterative robot–environment updates. However, the underlying scene representation degrades over time under repeated modifications. Splat-MOVER~\cite{Shorinwa2024_Splat-MOVER} introduces SEE-Splat, a scene-editing module that visualizes object motion through semantic masking and Gaussian-based infilling. While computationally efficient, interpolation-based infilling is limited in its ability to reconstruct occluded geometry and previously unseen regions. These limitations highlight the need for scene representations that support repeated object reconfiguration while preserving geometric and visual consistency. Motivated by this and by the importance of maintaining realistic remnants of removed or relocated objects~\cite{Kocour2025}, we propose a multiview-consistent diffusion-based inpainting approach that enables accurate scene editing and reliable future state prediction for robot–environment interaction.

\vspace{-2mm}
\section{METHODOLOGY}
    \label{sec:methodology}

\begin{figure}[t]
    \centering
    \includegraphics[width=0.99\columnwidth]{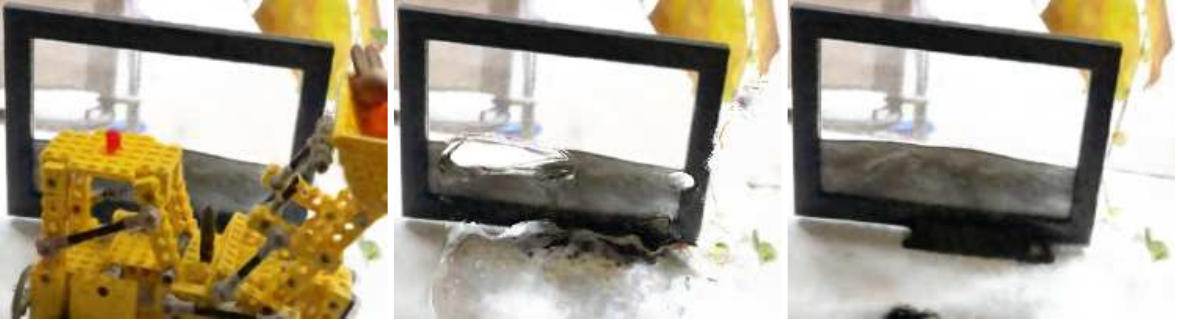}
    \put(-245,61){\color{white}a} \put(-164,60){\color{white}b} \put(-80,61){\color{white}c}
    \vspace{-2mm}
    \caption{Comparison of our neural field resampling method to the DFF~\cite{kobayashi2022decomposing} in the task of removing a yellow Lego dozer from the table surface: (a) table with object before editing operation, (b) the scene after
removing the dozer with DFF baseline~\cite{kobayashi2022decomposing}, and (c) our \our{} method.}
    \label{fig:sampling_res}
\end{figure}

The architecture of the proposed \our{} model is shown in Fig.~\ref{fig:architecture}, together with its integration in the robotic manipulation pipeline\footnote{The implementation is available at \url{https://csiro-robotics.github.io/NEO/}}. The network $F_{\Theta}$ is trained on a set of images $\mathcal{I} = \{ I_j \}_{j=1}^{N}$ captured by the robot in a single scan, together with the corresponding set of camera poses $\Omega = \{ \Omega_j \}_{j=1}^{N}$. The network learns a joint neural representation consisting of a scene field $\mathcal{S}\in\mathbb{R}^4$ and an associated language field $\mathcal{L}\in\mathbb{R}^{d_l}$, where ${d_l}$ is the dimensionality of the language feature vector.

\begin{equation}
(\mathcal{S}, \mathcal{L}) = F_{\Theta}(\mathcal{I}. \Omega),
\end{equation}

Given a user prompt $p$ describing the desired manipulation, we localize the target object region $\mathcal{O} \subset \mathbb{R}^3$ using the language field. Let $\phi(p) \in \mathbb{R}^{d_l}$ denote the text embedding of the prompt and $\mathcal{L}(\mathbf{x}) \in \mathbb{R}^{d_l}$ the language feature at a 3D surface point $\mathbf{x} \in \mathbb{R}^3$ extracted from $\mathcal{S}$. We compute the cosine similarity between $\phi(p)$ and $\mathcal{L}(\mathbf{x})$ over a set of such surface points to obtain a set of high-relevance points. These points are clustered, and the largest cluster is used to fit an oriented bounding box $b=(\mathbf{c},\theta)$, where $\mathbf{c} \in \mathbb{R}^3$ is the box center and $\theta \in \mathrm{SO}(2)$ denotes a planar rotation (yaw) about the vertical axis. This bounding box is used for object removal and for generating candidate grasps $G$.

\begin{figure}[t]
    \centering
    \includegraphics[width=0.79\columnwidth]{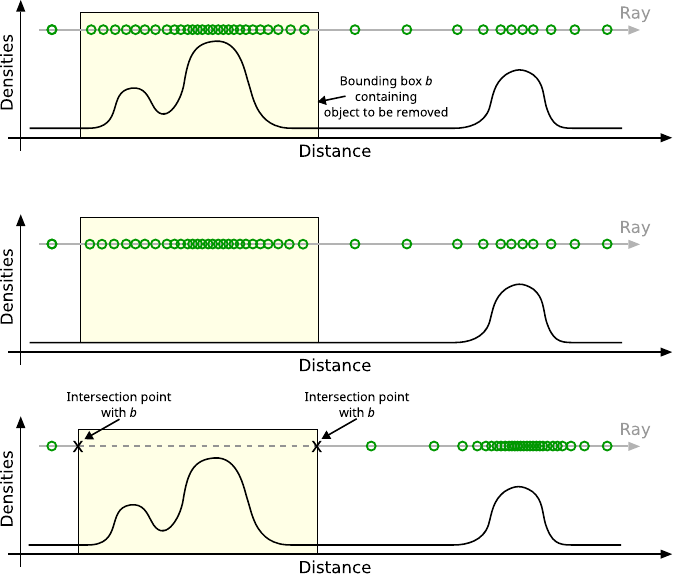}
    \put(-203,164){a} \put(-203,101){b} \put(-203,42){c}
    \vspace{-2mm}
    \caption{Proposed NeRF resampling method. The yellow color shows the region from which we would like to remove an object. Ray samples are represented by green circles. The top diagram depicts how the standard sampling procedure samples actual scene densities. The middle image is the most common practice of suppressing unwanted density values, such as done in DFF~\cite{kobayashi2022decomposing}. The bottom shows our resampling method, which avoids sampling the object region and focuses on other areas.}
    \label{fig:sampling}
\end{figure}

Conditioned on the prompt $p$ and a planned object motion $\tau$, \our{} produces an updated scene representation $(\mathcal{S}', \mathcal{L}')$ without requiring additional re-scanning:

\begin{equation}
(\mathcal{S}', \mathcal{L}') = \our\big((\mathcal{S}, \mathcal{L}),\, p,\, \tau\big).
\end{equation}

The updated representation can be reused as input to the same procedure for subsequent edits in sequential manipulation scenarios. For grasp execution, AnyGrasp~\cite{fang2023anygrasp} is used to generate candidate grasps, and the robot trajectory is planned using the most reliable grasp.

\subsection{Neural Field Resampling for Object Removal}
Building on the joint scene and language representation introduced above, we construct an editable \our{} model by first removing the user-specified target object from the NeRF.
Prior work on NeRF object removal (\eg{} DFF~\cite{kobayashi2022decomposing}) commonly edits densities along camera rays to suppress the density associated with the unwanted object. While this strategy is computationally efficient, it can introduce distortions in regions that were previously occluded by the removed object. These regions become undersampled during volumetric rendering, which may lead to hallucinated structures and geometric artifacts, asshown in Fig.~\ref{fig:sampling_res}b.

Our method takes as input camera poses and an oriented object bounding box $b$ defining the region to be removed in the NeRF world frame. For each camera ray, we compute its intersection with $b$ and exclude the corresponding ray segment from sampling. The remaining ray intervals (typically a near and far segment) are then concatenated into a single continuous 1D sampling domain. We draw a fixed number of uniform samples from it, which reallocates samples that would otherwise fall inside $b$ to the rest of the ray and mitigates under-sampling in newly revealed regions.

Next, we predict the densities at each sampled location and convert these densities to volumetric rendering weights using the standard NeRF formulation~\cite{mildenhall2020nerf}. These weights are then normalized to define a sampling distribution for hierarchical (PDF-based) resampling. We draw a second, fixed set of samples from this distribution over the same concatenated sampling domain, which increases sampling density in regions with a higher probability of occupied space while ensuring that no samples are drawn inside $b$. This two-stage sampling procedure enables immediate object removal and reduces artifacts behind the removed object. The proposed method generalizes to single or multiple bounding boxes in the scene.

Figure~\ref{fig:sampling}a illustrates the standard sampling of densities. The middle diagram (Fig.~\ref{fig:sampling}b) depicts DFF~\cite{kobayashi2022decomposing}, which can leave the background sparsely sampled. Figure~\ref{fig:sampling}c visualizes our proposed resampling method, where the object region is skipped, and sampling is concentrated on regions with a higher probability of occupied space.

The impact of this is shown in Fig.~\ref{fig:sampling_res}.
Fig.~\ref{fig:sampling_res}a shows the original NeRF rendering. Fig.~\ref{fig:sampling_res}b shows object removal with the baseline method~\cite{kobayashi2022decomposing}, where sparse sampling behind the removed object produces floating points and distorted regions. In contrast, our resampling yields a more consistent reconstruction of the revealed background (black frame and desk in Fig.~\ref{fig:sampling_res}c), which in turn provides a more reliable geometric representation for subsequent manipulation steps.

\subsection{Multiview-Consistent Progressive Inpainting}

\begin{figure}[t]
    \centering
    \includegraphics[width=0.99\columnwidth]{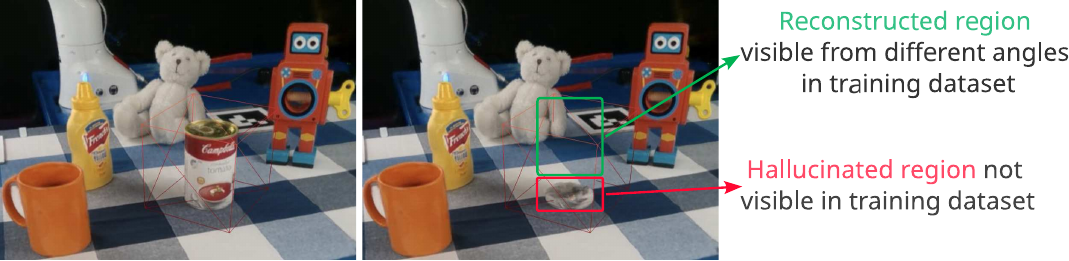}
    \put(-246,53){\color{white}a} \put(-163,52){\color{white}b}
    \vspace{-2mm}
    \caption{(a) Images of the scene before and (b) after removing the selected object (tomato soup can) containing a hallucinated region underneath the removed object, not visible in the training dataset.}
    \label{fig:inpainting_problem}
\end{figure}

After object removal, regions that were never observed in the original training views become weakly constrained and may be rendered with artifacts or implausible structure (\eg{} surfaces previously occluded underneath the ``tomato soup can'' in Fig.~\ref{fig:inpainting_problem}). To recover these missing regions, we propose a multiview-consistent progressive inpainting approach that (i) synthesizes additional supervision focused on the occluded area and (ii) integrates this supervision into NeRF training through explicit mask conditioning to jointly reconstruct missing content while preserving observed scene details.

\begin{figure}[t]
    \centering
    \includegraphics[width=0.99\columnwidth]{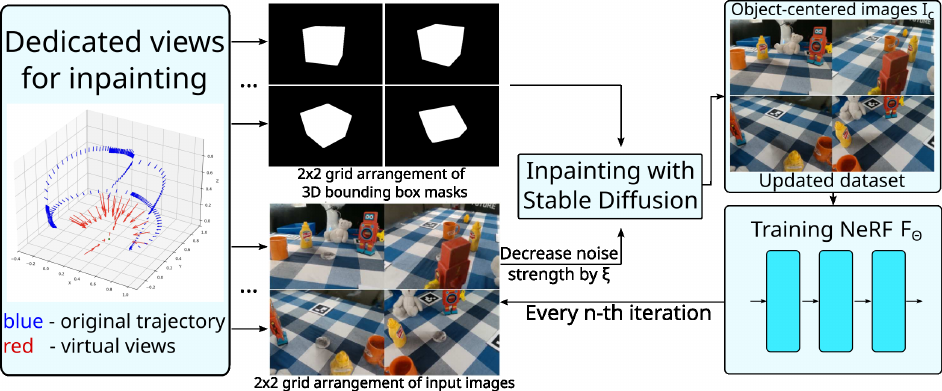}
    \vspace{-6mm}
    \caption{Multiview consistent progressive inpainting method and NeRF $\tilde{F_\Theta}$ training with removed object using iterative inpainting.}
    \label{fig:inpainting_training}
\end{figure}

An overview of the proposed inpainting pipeline is shown in Fig.~\ref{fig:inpainting_training}. Using the bounding box $b$, we augment the training set with additional virtual camera poses that focus on the local region around the removed object. Concretely, we define a virtual hemisphere centered at the bounding-box center and place virtual viewpoints on this hemisphere with optical axes oriented toward the center, at four opposing azimuths and varying elevations, forming a short upward spiral over the hemisphere (see Fig.~\ref{fig:inpainting_training}). This view augmentation concentrates supervision on surfaces that were previously occluded and improves cross-view coherence during subsequent inpainting and NeRF refinement.

We then build a training set $\mathcal{D}$ that combines (i) object-centered renderings from the virtual viewpoints, denoted $ I_v $, and (ii) the original captured images $I_o$, together with binary masks that identify the removed region. For each virtual view, we generate an object mask $M_v$ by projecting the 3D bounding box $b$ into the image plane and rasterizing the projected box faces to obtain a 2D mask; the complementary mask for original images is $M_o = 1 - M_v$.

Within each group, we tile four images into a single $2\times2$ composite and tile their corresponding masks in the same layout, inspired by NeRFiller~\cite{weber2023nerfiller}. A diffusion-based inpainting model is then conditioned on the composite mask, encouraging mutually consistent completions across views at similar elevations.
After inpainting, the composite is split back into individual views.
The inpainted object-centered virtual images $I_v$, together with the original images $I_o$ and their respective masks $(M_v, M_o)$, are used to train a NeRF that is explicitly conditioned on the masks.
This mask-conditioned training allows the model to selectively reconstruct regions previously occluded by the removed object while preserving the observed scene content outside the masked area.
Unlike NeRFiller, which inpaints the entire dataset and does not condition NeRF training on object masks, our approach performs inpainting only on a small set of dedicated virtual object-centered views and relies on mask-conditioned NeRF training to integrate the completed content into the scene.

\vspace{-2mm}
\begin{equation}
\mathcal{L}_{\mathrm{rec}} =
\mathbb{E}_{(I,M,\Omega)\sim\mathcal{D}}
\left[
\ell\!\left(\mathcal{R}(\tilde{F}_{\Theta}, \Omega),\, I;\, M\right)
\right],
\label{eq:lrec}
\end{equation}
where $\mathcal{R}(\tilde{F}_{\Theta}, \Omega)$ renders $\tilde{F}_{\Theta}$ from camera pose $\Omega$ associated with image $I$, and $\ell(\cdot,\cdot;M)$ is a mask-conditioned reconstruction loss that compares the rendered image to $I$ over pixels selected by $M$.
The training tuples $(I,M,\Omega)$ are sampled from $\mathcal{D}$, which includes the original captured views $(I_o,M_o,\Omega)$ and the inpainted virtual object-centered views $(I_v,M_v,\Omega)$. Here, $I$ is an RGB image, $M$ is a binary mask (with $M_v$ indicating the removed region and $M_o = 1 - M_v$).

The iterative training of the mask-conditioned NeRF $\tilde{F}_{\Theta}$ after object removal is shown in Fig.~\ref{fig:inpainting_training}. During optimization, the virtual object-centered views $I_v$ are re-rendered every $n$ steps, and the inpainting process described above is repeated to progressively refine the completed regions. After each inpainting iteration, we reduce the inpainting strength factor $\xi$, preserving more of the current rendering in later iterations and promoting convergence toward a stable reconstruction.

Next, to explicitly discourage density from reappearing inside the removed-object region, we introduce a regularization loss term that penalizes non-zero density predictions within the bounding box $b$.
\vspace{-4mm}

\begin{equation}
        \mathcal{L}_{reg} =\frac{1}{n}\sum_{i=1}^{n}\left( \tilde{\sigma}_i - \tilde{\sigma}'_i \right)^2,
        \label{eq: regularization_loss}
\end{equation}\vspace{-2mm}
where  \vspace{-2mm}

\begin{equation}
        \tilde{\sigma}' =
        \begin{cases}
            \tilde{\sigma}'(i) = 0, & \textbf{p}(i) \in b  \\[1ex]
            \tilde{\sigma}'(i) = \tilde{\sigma}(i), &  \textbf{p}(i) \notin b,
        \end{cases}
\end{equation}
$\tilde{\sigma}$ denotes the density values predicted by the $\tilde{F_\Theta}$, $\mathbf{p}$ represents 3D sample points, and $b$ is the bounding box of the object.

This loss, therefore, applies only to samples within $b$, reducing spurious density (“floaters”) and supporting persistent object removal.
Finally, we optimize $\tilde{F}_{\Theta}$ using the combined objective $\mathcal{L}_\text{masked}= \lambda_{rec}\mathcal{L}_{rec}+ \lambda_{reg}\mathcal{L}_{reg}$, where $\lambda_{rec}$ and $\lambda_{reg}$ are weighting constants.
At the end of this stage, we obtain a scene representation in which the object is removed, and the previously occluded regions are completed with improved geometric and visual consistency.

\subsection{Knowledge Distillation Method for Scene Reconfiguration}
\label{subsec:distillation}

\begin{figure}[t]
    \centering
    \includegraphics[width=0.99\columnwidth]{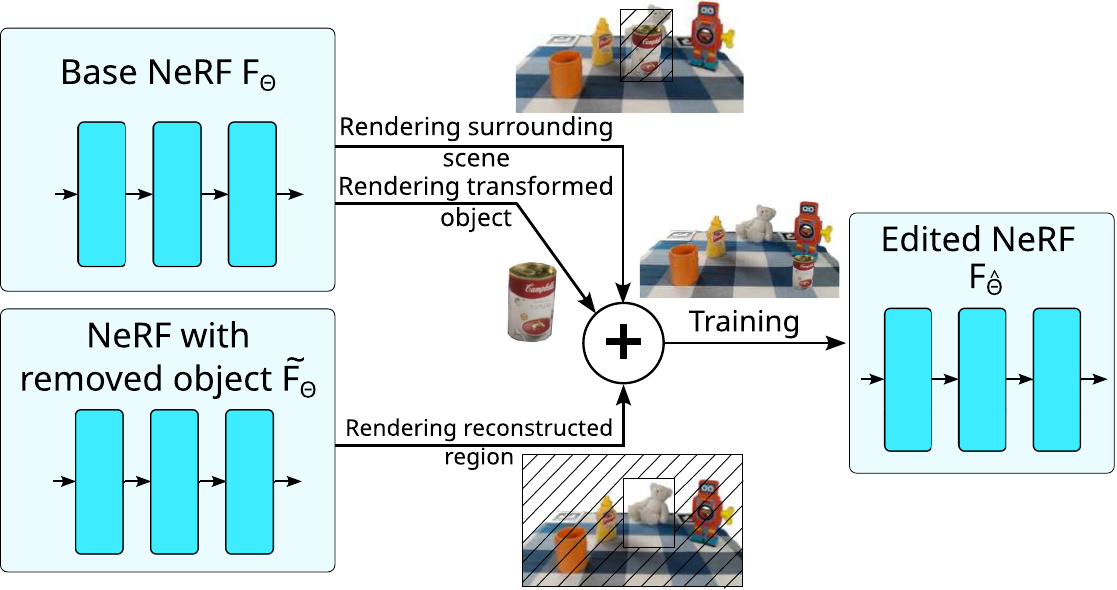}
    \vspace{-6mm}
    \caption{Object reconfiguration module for preservation of the structural integrity of the environment while reconfiguring objects using knowledge distillation via teacher-student networks.}
    \label{fig:reconfiguration}
\end{figure}

The previous stages produce two complementary scene representations: (i) the original NeRF $F_{\Theta}$, which encodes the complete scene including the target object, and (ii) an edited NeRF $\tilde{F}_{\Theta}$, in which the object has been removed and previously occluded regions have been reconstructed. The goal of this stage is to combine these representations into a single NeRF, $F_{\hat{\Theta}}$, that reflects the desired object reconfiguration while preserving global scene consistency (Fig.~\ref{fig:reconfiguration}).

    Our key idea is to construct a final scene representation $\mathcal{S}'$ by selecting supervision from multiple teacher networks rather than distilling from a single fully edited teacher like in Seal-3D~\cite{Wang2023_SEAL3D}.

    Specifically, rendering is decomposed into three disjoint regions: (i) outside the object bounding box supervised by $F_\Theta$, (ii) inside the removal bounding box supervised by $\tilde{F_\Theta}$, and (iii) regions corresponding to the relocated object supervised by $F_\Theta$ under the desired motion (Fig.~\ref{fig:reconfiguration}).

    This region-wise composition defines a target field $\mathcal{S}'$, which is used to supervise a student NeRF $F_{\hat{\Theta}}$ via volumetric rendering. The student parameters $\hat{\Theta}$ are initialized from $\Theta$, while the language field parameters are fixed. The student is optimized to minimize discrepancies between rendered color, density, and depth values induced by $\mathcal{S}'$.
    To focus supervision on modified regions of the scene, training additionally incorporates \ccyan{virtual} object-centered views \ccyan{$I_v$} introduced during the inpainting stage.
    These views increase ray density in areas affected by object removal and relocation, improving the fidelity of the distilled representation.

    Knowledge distillation is performed in two stages. In the initial phase, the student is supervised directly using teacher-predicted color and density values along rays. In the second phase, training proceeds using a combination of the volumetric RGB reconstruction loss plus depth to match rendered RGB and depth images from teacher. Based on this formulation, we define the distillation loss $\mathcal{L}_{\mathrm{distill}}$ as follows:

\vspace{-2mm}
\begin{equation}
\mathcal{L}_{\mathrm{distill}} =
\begin{cases}
\displaystyle \frac{1}{rm}\sum_{j=1}^{r}\sum_{k=1}^{m}
\Big(\,|\sigma_{j,k}-\hat{\sigma}_{j,k}|+
\|c_{j,k}-\hat{c}_{j,k}\|_1\,\Big) \\[0.5ex]
\displaystyle \frac{1}{n}\sum_{i=1}^{n}
\Big(\,\|C_i-\hat{C}_i\|_2+\|D_i-\hat{D}_i\|_2\,\Big),
\label{eq:distill_loss}
\end{cases}
\end{equation}

where $m$ is the number of samples per ray, $r$ is the number of rays, and $n$ is the number of pixels. The indices $j$, $k$, and $i$ denote the ray index, the sample index along a ray, and the pixel index, respectively. $\sigma_{j,k}$ and $c_{j,k}$ are the teacher density and color at sample $k$ on ray $j$, while $\hat{\sigma}_{j,k}$ and $\hat{c}_{j,k}$ are the corresponding student predictions. $C_i$ and $D_i$ denote the teacher rendered RGB and depth at pixel $i$, while $\hat{C}_i$ and $\hat{D}_i$ denote the corresponding student renderings.

    The resulting parameters $\hat{\Theta}$ define an updated NeRF that encodes the reconfigured scene $\mathcal{S}'$, enabling novel-view rendering with consistent object relocation, complete object removal, and plausible reconstruction of previously occluded regions.

\subsection{Updating Language Fields}

After geometric editing, we keep the edited NeRF \ccyan{$F_{\hat{\Theta}}$} fixed and update the associated language field \ccyan{$\mathcal{L}'$} so that it remains consistent with the modified scene and supports subsequent text queries.
Supervision for updating $\mathcal{L}$ comes from CLIP and DINO features extracted from \ccyan{new images rendered from $F_{\hat{\Theta}}$ at the original camera poses $\Omega$}.
In practice, the optimization of \ccyan{$\mathcal{L'}$} converges in a small number of iterations. We hypothesize that this is because CLIP and DINO features primarily capture high-level semantic content and are less sensitive to fine geometric detail, allowing the dominant semantic structure to be recovered early during training.

\subsection{Editing Many Times}

    The proposed \our{} model supports iterative scene editing, which is essential for multi-step robotic manipulation and assembly tasks. After each editing operation, the resulting scene representation, consisting of a NeRF and an associated language field, retains the same structure as the original input representation. As a result, the edited model can be directly reused as input for subsequent editing steps. Across iterations, regions of the scene unaffected by the current edit are reconstructed using the original NeRF representation, preserving visual and geometric consistency while enabling progressive object-level modifications.

\vspace{-3mm}
\section{DATASET}
\label{sec:dataset}

    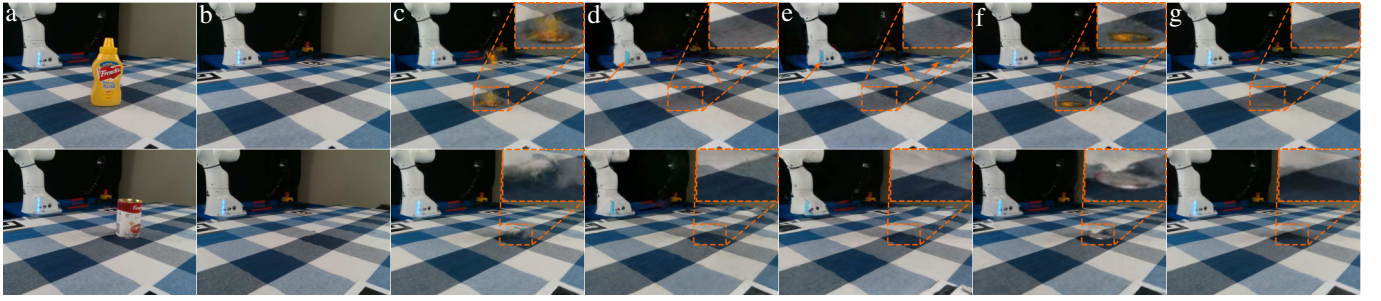
\begin{figure*}[t]
        \centering
        \includegraphics[width=0.99\textwidth]{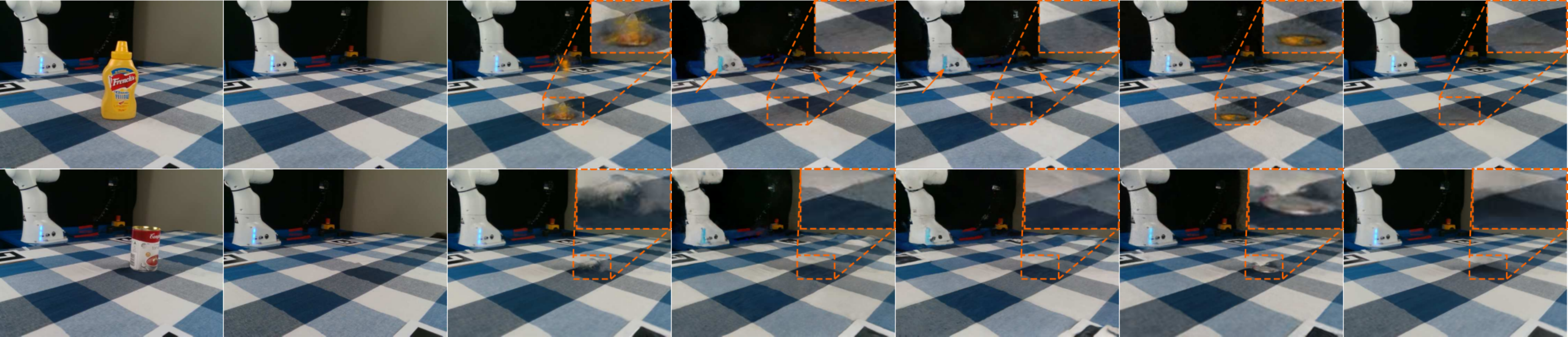}
        \put(-510,103){\color{white}a} \put(-437,102){\color{white}b} \put(-364,103){\color{white}c} \put(-291,102){\color{white}d} \put(-218,103){\color{white}e} \put(-145,101){\color{white}f}
        \put(-72,103){\color{white}g}
        \vspace{-2mm}
        \caption{Example scenes after removing the selected object: (a) original scenes with object, (b) ground truth images after object removal, (c) DFF~\cite{kobayashi2022decomposing}, (d) NeRFiller~\cite{weber2023nerfiller}, (e) NeRFiller${}_{\rm w/o D}$~\cite{weber2023nerfiller}, (f) Seal-3D~\cite{Wang2023_SEAL3D}, and (g) \our{} (ours). Regions corresponding to the areas after object removal are shown in enlarged views, and the deformations introduced by NeRFiller are indicated with arrows.}
        \label{fig:removal_rgb_selected}
    \end{figure*}

A critical limiting factor in assessing NeRF editing models for use in real-world robotic applications is the lack of benchmarks that enable quantitative evaluation of scene edits against real-world changes within a scene.
To address this gap, we present the \textit{\our-Dataset} benchmark which provides ground-truth, registered tabletop scenes captured both before and after object manipulation.

\textit{\our-Dataset} comprises twelve total tabletop scenes used to evaluate NeRF editing for object removal and manipulation edits.
This includes 11 single-object scenes and 1 multi-object scene.
The scenes were captured using two different robotic setups with different fixed robotic arms (Franka Emika Panda and UR5) and tabletops to introduce variability in the benchmark under controlled illumination, avoiding strong specular and lighting effects.
Each scene $\mathcal{S}$ is scanned before and after object manipulations using an automated capture procedure in which the robot arm moves a camera along a fixed trajectory while acquiring images, ensuring consistent scene coverage across scans.
An example of the camera frames produced by this process can be seen on the left of Fig.~\ref{fig:inpainting_training}, colored in blue. The captured images are used both to train the initial model $F_\Theta$ and as ground truth for evaluating rendered results after editing. For registration purposes, we placed four ArUco markers on the table surface. These markers are detected in the scans and used to compute a rigid transformation that aligns the scenes before and after object manipulations, recovering the scale, orientation, and origin of the scene in the world coordinate frame. ArUco markers are not used for training.

The \textit{\our-Dataset} includes both object removal and object manipulation scenarios for evaluating NeRF editing methods. For object removal, scans are recorded before and after the object(s) are removed. For object manipulation, the scene is first scanned and the object(s) are then moved by a predefined transform relative to their initial pose, using the same fixed robotic arms to ensure consistent execution. The manipulated object classes are drawn from the YCB dataset~\cite{calli2015benchmarking}.

\vspace{-3mm}
\section{EXPERIMENTAL DESIGN}

We compare \our{} against several recent NeRF editing SOTA baselines. DFF~\cite{kobayashi2022decomposing} introduced NeRF editing and supports object removal, Nerfiller~\cite{weber2023nerfiller} is designed for inpainting missing regions in neural radiance fields, and Seal-3D~\cite{Wang2023_SEAL3D} is currently the only method capable of moving objects while updating NeRF weights. For evaluation, we employ standard NeRF quality (PSNR, SSIM) and geometry metrics (depth RMSE and photometric reprojection error $E_{\mathrm{rep}}$ \cite{Engel2014} that measure 3D consistency of the obtained models) computed over both the full image and the edited regions, allowing us to assess both unedited fidelity and reconstruction quality.

\subsection{Object Removal Experiments}
To evaluate the effectiveness of \our{} for object removal, we use five single-object scenes from the \textit{\our{}-Dataset} in which a single object has been physically removed between scans. The evaluation is computed at two levels of spatial focus, full image and masked region. For full-image evaluation, the rendered output from the edited NeRF is compared against the entire ground-truth image. A drop in these metrics can indicate that the model has lost fidelity in unedited regions of the scene, reflecting potential catastrophic forgetting due to the editing process. In the masked-region evaluation, metrics are computed only within the object removal region defined by the bounding box described in Section~\ref{sec:methodology}. For simplicity, we refer to this region as "Out", indicating that the object has been moved out of it. These measurements isolate the model’s ability to plausibly reconstruct previously occluded regions and match them to real-world appearance.

\subsection{Robotic Manipulation Experiments}

To assess the ability of \our{} to capture the outcome of real-world object manipulations, we use six scenes from the \textit{\our{}-Dataset}, each with a single object moved by a robot to a known target pose. Following the same evaluation protocol as in the object removal experiments, we assess editing performance using the standard NeRF metrics computed over two masked regions: ``Out'', corresponding to the object's original location prior to movement, and ``In'', corresponding to its new location after the manipulation. These region-based metrics allow us to separately assess the model’s ability to reconstruct newly revealed surfaces (Out) and to render the relocated object at its new position (In), in alignment with real-world observations.

\subsection{Sequential Manipulation Experiments (Assembly)}
To demonstrate the use of \our{} for consecutive edits and downstream robotic tasks, we conduct an assembly experiment in which a sequence of manipulation actions is planned and executed using our model as the scene representation.
The objective is to assemble multiple objects into a target configuration through a series of iterative edits and robot actions. We begin by constructing an initial scene model and then perform four consecutive manipulation steps. At each step, interactive language prompts are used to identify the target object and specify its desired motion, expressed as a relative 6-DoF transformation.
For each manipulation, \our{} predicts the resulting scene state prior to physical execution, allowing the user to visualize the expected outcome. Once confirmed, the robot autonomously moves the selected object to the specified location without further human intervention.

Both bounding box generation and grasp planning are performed directly on the current edited \our{} model, which incorporates updated geometry and a dynamically refined language field. This experiment highlights \our{}’s ability to not only render the visual result of edits, but also to maintain a consistent semantic representation across sequential manipulations, enabling it to serve as a live state estimator for manipulation planning.

For the manipulation sequence, only the initial scene and reference object motions are provided in \textit{\our-Dataset}, requiring intermediate states to be produced by the editing method and enabling evaluation of the full scene update pipeline.

\vspace{-2mm}
\section{RESULTS}

\subsection{Objects Removal Results}

\begin{table}[t]
    \centering
    \caption{Average metrics calculated across five different scenes in the object removal scenario.}\vspace{-3mm}
    \scriptsize
    \small
    \vspace{3pt}
    \setlength{\tabcolsep}{0.7pt}
    \renewcommand{\arraystretch}{0.9}
    \begin{tabular}{lllllllllll}
        \toprule
         \multirow{2}{*}{Method} &
         \multicolumn{2}{c}{PSNR [dB] $\uparrow$} &
         \multicolumn{2}{c}{SSIM [-] $\uparrow$} &
        \multicolumn{2}{c}{RMSE [m] $\downarrow$} &
        \multicolumn{2}{c} {${E}_{\rm rep}$ [-] $\downarrow$} \\
        \cmidrule(lr){2-3} \cmidrule(lr){4-5} \cmidrule(lr){6-7} \cmidrule(lr){8-9}
         & Full & Out & Full & Out & Full & Out & Full & Out \\
        \midrule

        NeRFiller~\cite{weber2023nerfiller} & \cc{20} 20.40 & \cc{60} 21.71 & \cc{20} 0.696 & \cc{20} 0.671 & \cc{20} 0.252 & \cc{20} 0.254 & \cc{20} 0.104 & \cc{60} 0.080\\

        Seal-3D \cite{Wang2023_SEAL3D} & \cc{60} 27.08 & \cc{20} 20.05 & \cc{80} 0.866 & \cc{40} 0.709 & \cc{40} 0.162 & \cc{40} 0.140 & \cc{40} 0.079 & \cc{20} 0.089 \\

        $\text{NeRFiller}_{\text{w/o D}}$ & \cc{40} 22.19 & \cc{80} 23.73 & \cc{40} 0.748 & \cc{60} 0.729 & \cc{60} 0.080 & \cc{60} 0.096 & \cc{80} 0.076 & \cc{60} 0.080 \\

        DFF~\cite{kobayashi2022decomposing} & \cc{80} 26.91 & \cc{40} 21.64 & \cc{100} 0.878 & \cc{80} 0.747 & \cc{100} \bf 0.048 & \cc{80} 0.047 & \cc{80} 0.068 & \cc{80} 0.074 \\

        \our{}~(Ours) & \cc{100} \bf 27.20 & \cc{100} \bf 25.43 & \cc{100} \bf 0.881 & \cc{100} \bf 0.845 & \cc{80} 0.056 & \cc{100} \bf 0.033 & \cc{100} \bf 0.067 & \cc{100} \bf 0.071 \\
        \bottomrule
    \end{tabular}
    \label{tab:cv_results}
\end{table}

Tab.~\ref{tab:cv_results} reports the quantitative results for object removal.
We observe that, in this task, \our{} consistently outperforms two variants of NeRFiller~\cite{weber2023nerfiller} (with and without depth supervision), Seal-3D~\cite{Wang2023_SEAL3D}, and the DFF baseline~\cite{kobayashi2022decomposing}, in both full-image and masked-image evaluation settings. In general, \our{} achieves the best performance across all metrics.
A qualitative comparison, shown in Fig.~\ref{fig:removal_rgb_selected}, is consistent with these quantitative results. Our method produces images that closely resemble the ground-truth scene. By contrast, DFF (Fig.~\ref{fig:removal_rgb_selected}c) leaves floating artifacts after object removal. NeRFiller (Fig.~\ref{fig:removal_rgb_selected}d) and NeRFiller$_{\text{w/o D}}$ reconstruct locally plausible regions but introduce more background artifacts, while Seal-3D (Fig.~\ref{fig:removal_rgb_selected}f) leaves many artifacts on the table surface.
In addition to these accuracy improvements, \our{} is computationally efficient, completing scene inpainting in approximately 6 minutes, compared to 13 minutes for DFF and about 1.5 hours for NeRFiller, resulting in an approximately 15$\times$ speed-up.

\subsection{Robotic Manipulation Results}

    \begin{figure}[t]
        \centering
        \includegraphics[width=1.0\columnwidth]{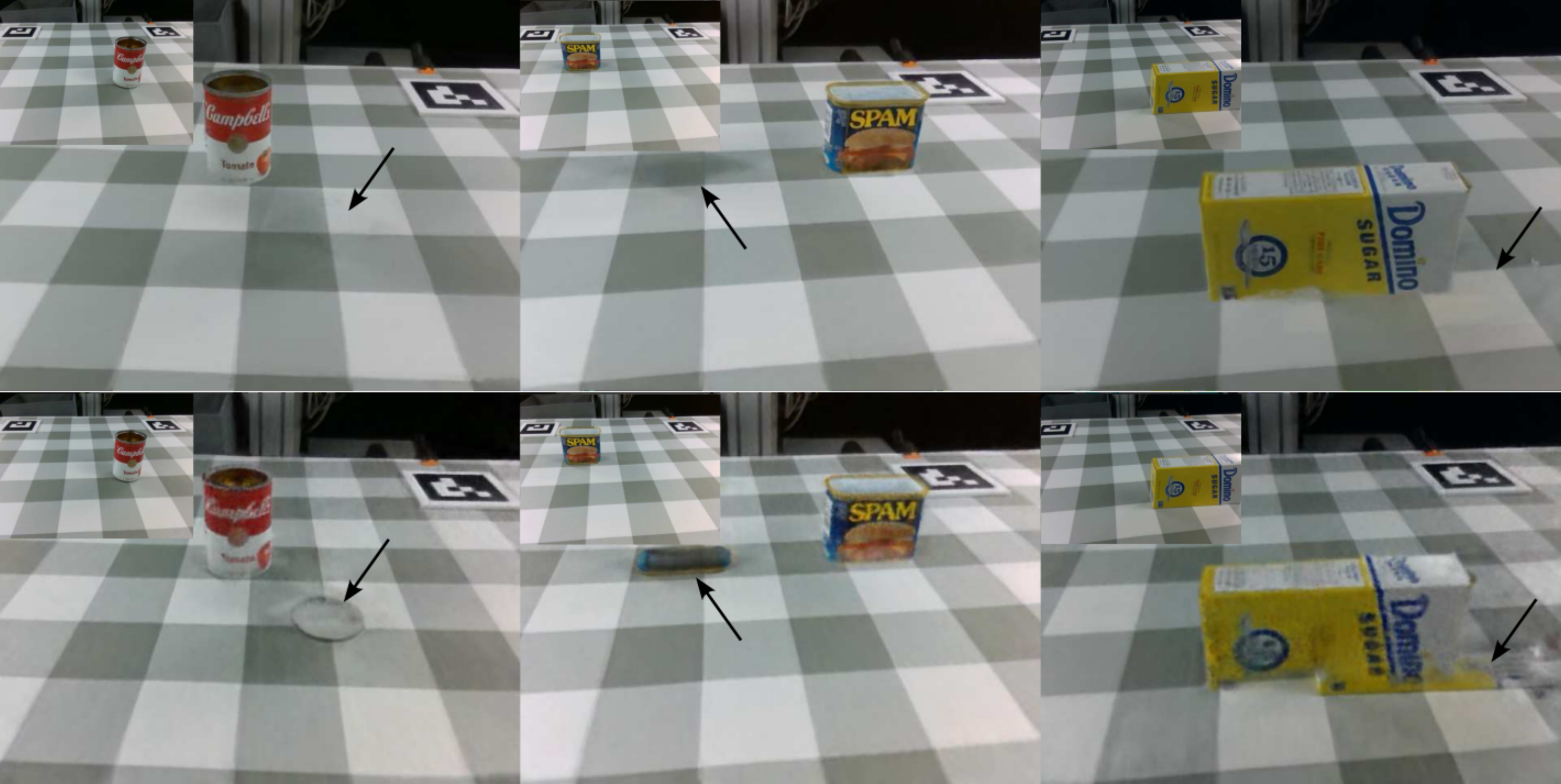}
        \put(-251,120){\color{white}a} \put(-251,55){\color{white}b}
        \caption{Example images generated by (a)  \our{} and (b) Seal-3D~\cite{Wang2023_SEAL3D} after object manipulation experiments. Corner images and arrows show the initial configurations of objects and artifacts generated by both methods, respectively.}
        \label{fig:pick_and_place_results}
    \end{figure}

We compare \our{} against Seal-3D~\cite{Wang2023_SEAL3D} in a pick-and-place scenario. Qualitative results are presented in Fig.~\ref{fig:pick_and_place_results}, demonstrating that \our{} produces higher-quality reconstructions of moved objects. Seal-3D often leaves parts of the original object at its previous location and shows noticeable degradation in the appearance of the manipulated object. In contrast, \our{} avoids these visual artifacts. These observations are supported by the quantitative results in Tab.~\ref{tab:pick_and_place_results} where \our{} consistently achieves the best photometric scores, depth RMSE, and reprojection error $E_{\mathrm{rep}}$. \our{} completes the object manipulation task in about 1.5 minutes, compared to 3 minutes for Seal-3D, yielding a 2$\times$ speed-up due to more efficient teacher–student knowledge distillation.

\begin{table}[t]
    \centering
    \caption{Results of pick and place experiments showing the metrics between the images generated by \our{} and ground-truth images from the scene after moving the object.}
    \vspace{-2mm}
    {\footnotesize
    \setlength{\tabcolsep}{3.5pt}
    \renewcommand{\arraystretch}{0.9}
    \begin{tabular}{ccccccccc}
        \toprule
        \multirow{2}{*}{Method} &
        \multicolumn{2}{c}{PSNR [dB] $\uparrow$} &
        \multicolumn{2}{c}{SSIM [-] $\uparrow$} &
        \multicolumn{2}{c}{RMSE [m] $\downarrow$} &
        \multicolumn{2}{c}{$E_{\rm rep}$ [-] $\downarrow$}\\
        \cmidrule(lr){2-3} \cmidrule(lr){4-5} \cmidrule(lr){6-7} \cmidrule(lr){8-9}
        & Out & In & Out & In & Out & In & Out & In\\
        \midrule
Seal-3D \cite{Wang2023_SEAL3D} & 18.94 & 18.56 & 0.695 & 0.589 & 0.156 & \bf 0.182 & 0.102 & 0.118 \\
\our & \bf 21.99 & \bf 19.26 & \bf 0.783
& \bf 0.619 & \bf 0.141 & \bf 0.182 & \bf 0.046 & \bf 0.080\\

\bottomrule

    \end{tabular}
    }
    \label{tab:pick_and_place_results}
    \end{table}

\subsection{Ablation Study}

\begin{table}[t]
    \centering
    \vspace{-3mm}
    \caption{Ablation study on the pick and place task.}
    {\footnotesize
    \setlength{\tabcolsep}{1.55pt}
    \renewcommand{\arraystretch}{0.9}
    \begin{tabular}{lcccccccccc}
        \toprule
        \multirow{2}{*}{Method} &
        \multicolumn{2}{c}{PSNR [dB] $\uparrow$} &
        \multicolumn{2}{c}{SSIM [-] $\uparrow$} & \multicolumn{2}{c}{RMSE [m] $\downarrow$} & \multicolumn{2}{c}{$E_{\rm rep}$ [-] $\downarrow$}\\
        \cmidrule(lr){2-3} \cmidrule(lr){4-5} \cmidrule(lr){6-7} \cmidrule(lr){8-9}
        & Out & In & Out & In & Out & In & Out & In\\
        \midrule
        w/o Virtual view dist. & 17.34 & 18.89 & 0.637 & 0.614 & 0.134 & 0.169 & 0.054 & \bf 0.080 \\
        w/o Grid inp. & 15.83 & 18.42 & 0.590 & 0.600 & 0.142 & 0.173 & 0.062 & 0.081 \\
        w/o Mask-cond. train. & 13.09 & 18.65 & 0.502 & 0.605 & 0.157 & 0.171 & 0.088 & 0.083 \\
        Field distill only & 15.24 & 12.29 & 0.402 & 0.387 & 0.261 & 0.199 & 0.060 & 0.115 \\
        \our{} full model & \bf 21.99 & \bf 19.26 & \bf 0.783 & \bf 0.619 & \bf 0.056 & \bf 0.033 & \bf 0.046 & \bf 0.080\\

        \bottomrule
    \end{tabular}
    \label{tab:ablation_study}
    \vspace{-5mm}
    }
\end{table}

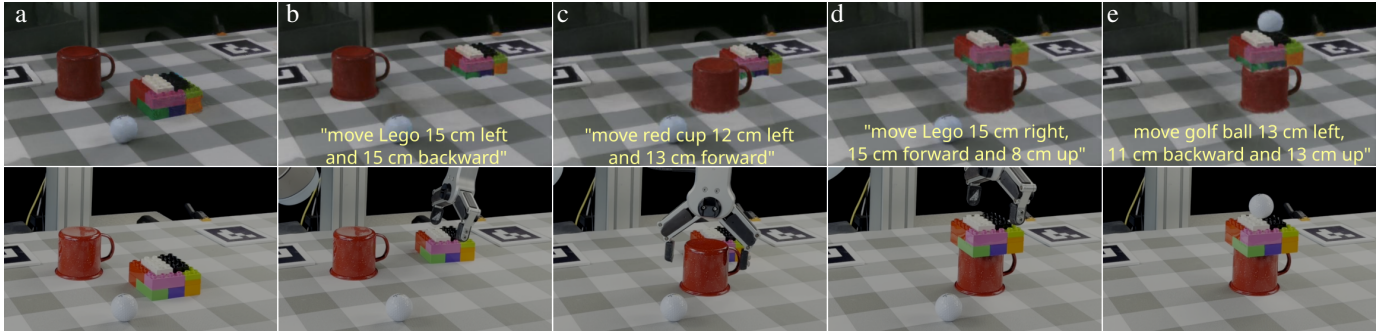
\begin{figure*}[t]
        \centering
        \includegraphics[width=1.0\textwidth]{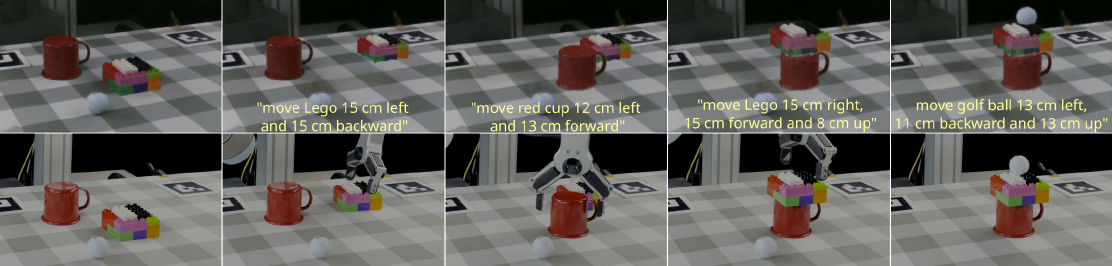}
        \put(-512,116){\color{white}a} \put(-410,116){\color{white}b} \put(-308,116){\color{white}c} \put(-205,116){\color{white}d} \put(-101,116){\color{white}e}
        \vspace{-3mm}
        \caption{Assembly experiment showing \our{} output (top row), which predicts the future state of the scene based on the initial scene (a) and a series of text prompts, along with the ground-truth results of the real-robot experiment using the NEO model for object localization and grasp planning.}
        \label{fig:assembly}
        \vspace{-1mm}
\end{figure*}

In the ablation study, we compare the full \our{} model with variants removing the virtual object-centered view distribution (w/o virtual view dist. in Tab.~\ref{tab:ablation_study}), the $2\times2$ grid inpainting strategy (w/o grid inp.), and the mask-conditioned NeRF training objective (w/o mask-cond. train.). We also ablate distillation by training the student $F_{\hat{\Theta}}$ using only the first term of $\mathcal{L}_{\mathrm{distill}}$ in Eq.~(\ref{eq:distill_loss}) (field dist. only), enforcing consistency of teacher and student densities and colors along rays. The results in Tab.~\ref{tab:ablation_study} show that mask-conditioned training is most important, while virtual view distribution has a smaller impact and grid inpainting lies in between. Ablating $\mathcal{L}_{\mathrm{distill}}$ shows that RGB and depth supervision from student rendering is most valuable, and all components contribute to \our{} performance.

\subsection{Assembly Results}

Qualitative comparisons of our consecutive assembly experiment are shown in Fig.~\ref{fig:assembly}.  These show only small differences between the synthesized view and the real images of the scene during manipulation. Despite multiple actions, the synthesized views closely match real images, demonstrating that even after an edit, the \our{} model supports subsequent edits and accurately predicts the effects of each action.

\section{CONCLUSIONS}

In this paper, we present a unified framework for realistic, language-guided NeRF editing tailored to robotic manipulation. By combining object removal, knowledge distillation, and direct NeRF weight editing, our approach enables robots to anticipate the visual outcomes of interactions before execution. Through real-world pick-and-place and assembly experiments, we demonstrate that edited NeRFs reliably predict manipulation-induced scene changes. Across object removal and relocation scenarios, our method consistently outperforms prior NeRF editing approaches in terms of visual fidelity and structural consistency. These results show that editable NeRF representations can accurately reflect real-world scene changes and serve as reliable predictors for downstream manipulation tasks. We additionally provide a ground-truth benchmark dataset to facilitate future evaluation of NeRF editing methods. Future work will extend the framework to articulated objects and leverage \our{} to predict future scene states for training an RL-based controller in multi-stage assembly.

\addtolength{\textheight}{-12cm}

\bibliographystyle{IEEEtran}
\vspace{-4mm}
\bibliography{bibliography}

\end{document}